\documentclass{article}
\usepackage{ijcai25}

\usepackage{times}
\usepackage{soul}
\usepackage{url}
\usepackage[hidelinks]{hyperref}
\usepackage[utf8]{inputenc}
\usepackage[small]{caption}
\usepackage{graphicx}
\usepackage{amsmath}
\usepackage{amsthm}
\usepackage{booktabs}
\usepackage{algorithm}
\usepackage{algorithmic}
\usepackage[switch]{lineno}

\usepackage{amssymb}
\usepackage{bbm}
\usepackage{float}

\title{Adjusting the Output of Decision Transformer with Action Gradient}

\author{
Rui Lin$^1$
\and
Yiwen Zhang$^1$\and
Zhicheng Peng$^{2}$\And
Minghao Lyu$^1$\\
\affiliations
$^1$South China University of Technology\\
$^2$Sun Yat-sen University\\
}

\begin{document}

\maketitle

\newcommand{\bhline}{\noalign{\global\arrayrulewidth=1.2pt}\hline\noalign{\global\arrayrulewidth=0.4pt}}

\begin{abstract}

Decision Transformer (DT), which integrates reinforcement learning (RL) with the transformer model, introduces a novel approach to offline RL. Unlike classical algorithms that take maximizing cumulative discounted rewards as objective, DT instead maximizes the likelihood of actions. This paradigm shift, however, presents two key challenges: stitching trajectories and extrapolation of action. Existing methods, such as substituting specific tokens with predictive values and integrating the Policy Gradient (PG) method, address these challenges individually but fail to improve performance stably when combined due to inherent instability. To address this, we propose Action Gradient (AG), an innovative methodology that directly adjusts actions to fulfill a function analogous to that of PG, while also facilitating efficient integration with token prediction techniques. AG utilizes the gradient of the Q-value with respect to the action to optimize the action. The empirical results demonstrate that our method can significantly enhance the performance of DT-based algorithms, with some results achieving state-of-the-art levels.

\end{abstract}

\section{Introduction}

Reinforcement Learning (RL) has been effectively applied to various control tasks. However, challenges arise in specific domains, such as diagnostics and dialogue systems, where the agent cannot interact with a simulated environment ~\cite{levine2020offline}. In these scenarios, leveraging previously collected data for agent training becomes necessary. Furthermore, in complex tasks, online RL algorithms demonstrate suboptimal performance due to the limitations of random exploration in the early stage of training, which often fails to find learnable trajectories. Hence, offline RL has attracted widespread attention in recent years.

In addition to refining algorithms that demonstrate strong performance in online RL, another cutting-edge approach involves the integration of RL with the transformer model, as proposed by ~\cite{vaswani2017attention}. This model is characterized by its significant capacity for in-context learning ~\cite{radford2019language,brown2020language,akyurek2022learning,garg2022can}. The introduction of the Decision Transformer (DT) ~\cite{chen2021decision} illustrates the feasibility of moving beyond traditional algorithmic frameworks, enabling these powerful and rapidly evolving transformer-based models in RL applications.

The foundational distinction between DT-based algorithms and traditional RL algorithms leads to challenges in achieving superior performance. While traditional algorithms set the maximization of cumulative discounted rewards as goal ~\cite{sutton2018reinforcement}, the focus of  DT on maximizing the likelihood of actions conditioned on specific information results in extrapolation disadvantages, which can be classified into two categories: trajectory-level extrapolation, often referred to as stitching, and state-level extrapolation, which is the model's capability to infer actions that exceed those present in the dataset for a given state.

Multiple methodologies have been proposed to tackle the challenges associated with extrapolation. Despite certain unique methodologies ~\cite{janner2021offlinereinforcementlearningbig,hu2023graph,xie2023future,wang2024critic,huang2024context}, the remaining methods can be categorized into two main approaches: substituting the return-to-go with values predicted by alternative models (Token Prediction, TP) ~\cite{yamagata2023q,correia2023hierarchical,ma2023rethinking,wu2024elastic,zhuang2024reinformer}, and incorporating a policy gradient loss term into the loss function (PG) ~\cite{hu2024q,yan2024reinforcement}. This study investigates how the former approach enhances trajectory-level extrapolation, while the latter improves state-level extrapolation. However, the integration of these two approaches cannot stably yield satisfactory results due to the deadly triad ~\cite{vanhasselt2018deepreinforcementlearningdeadly}. To address this issue, we propose Action Gradient (AG), a framework designed to enhance state-level extrapolation abilities that can be conventionally integrated with TP.

An intuitive interpretation of AG is that it initially derives an action using DT. Subsequently, the trained critic is employed to conduct a heuristic search in the vicinity of this action to identify a refined action, which is then chosen for interaction with the environment. This approach is straightforward to implement and only requires modifications to the evaluation part, yet it significantly improves the extrapolation ability and enhances the algorithm's overall performance.

We conducted experiments on Gym and Maze2d datasets from the D4RL benchmark ~\cite{fu2020d4rl} to validate the effectiveness of AG. Through comparative analysis of theoretical and experimental results, we demonstrated the limitations of PG and the advantages AG offers. We also identified potential for future research focused on further optimizing the algorithm. These findings provide new insights and perspectives on integrating RL with transformer models.

\section{Background}

\subsection{Offline Reinforcement learning}

Reinforcement learning (RL) models the sequential decision problem as a Markov Decision Process (MDP), defined by $M=\{\mathcal{S}, \mathcal{A}, P, r, \gamma\}$. At each time step $t$, an agent observes the current state $s_t \in \mathcal{S}$ and selects an action $a_t \in \mathcal{A}$ based on a policy $\pi(a_t | s_t)$. The agent then receives a reward $r_t = r(s_t, a_t)$ and reaches a new state $s_{t+1} \in \mathcal{S}$ according to the state transition probability $P(s_{t+1} | s_t, a_t)$. We define a trajectory as $\tau = (s_0, a_0, r_0, \dots, s_T, a_T, r_T)$. The goal is to find a policy $\pi$ that maximizes the expected cumulative reward $J(\pi)=\mathbb{E}_{s_{t}\sim p_{\pi},a_{t}\sim\pi}\left[\sum_{t=0}^{T} \gamma^t r(s_t, a_t)\right]$.

In the context of offline RL, interaction with the environment is prohibited; instead, the agent relies solely on a fixed offline dataset $\mathcal{D}=\{\tau_i\}_i^{n-1}$ ~\cite{levine2020offline}. These trajectory data are generated through interactions between one or more unknown policies and the environment. 

\subsection{Decision Transformer}

Decision Transformer (DT) ~\cite{chen2021decision} presents a novel framework in which, at time step $t$, the model uses a preceding sequence of context length $k$ represented as $(s_{t-k}, RTG_{t-k}, a_{t-k}, \dots, s_t, RTG_t)$ to predict action $a_t$. Here, $RTG_{t} = \sum_{i=t}^{T}r_i$ denotes the return-to-go, enabling the model to make action predictions that are informed by future desired returns. During the evaluation phase, a preset $RTG_0$ is employed, and the return-to-go is subsequently updated according to the relation $RTG_{t} = RTG_{t-1} - r_{t-1}$.

In contrast to the manual selection and subsequent updating of the return-to-go value, employing a neural network to predict a value with similar properties presents a more practical approach. The paradigm of these methods is close to hierarchical RL ~\cite{nachum2018dataefficienthierarchicalreinforcementlearning}. Typically, Autotuned Decision Transformer (ADT) ~\cite{ma2023rethinking} utilizes Q-values and V-values that are trained through the Implicit Q-Learning (IQL) ~\cite{kostrikov2021offline}, while Reinformer ~\cite{zhuang2024reinformer} trains a model utilizing expectile regression to estimate the return-to-go value. These algorithms also involve other improvements to achieve good results. In the subsequent sections, TP only refers to the replacement of the presetting $RTG$ with a single predicted value.

An alternative approach to improving performance involves incorporating a policy gradient loss term into the loss function. This strategy aims to equip the model with the capability to extrapolate effectively. This method has demonstrated success in both the Offline RL domain ~\cite{hu2024q} and the Offline-to-Online RL domain ~\cite{yan2024reinforcement}. Notably, CGDT ~\cite{wang2024critic} also involves a critic network when optimizing the transformer model, but it differs in form from conventional methods. In this study, methods of this nature, as well as those akin to AWR ~\cite{wang2018exponentially,peng2019advantage,nair2020awac}, are not classified under the category of PG.

\section{Methodology}

\subsection{Extrapolation Ability}\label{sec:extrapolation}

The likelihood-based approach aims to imitate the behavior $\beta$ that generates the dataset rather than choosing actions with high expected returns. Consequently, it is inherently limited in its ability to outperform $\beta$. Although conditioning on return may lead the behavior performed by the agent closer to the trajectories with higher returns, employing trajectory-level information can compromise its stitching ability. This limitation of DT has been theoretically analyzed in previous research ~\cite{brandfonbrener2022does}.

\begin{figure}[t]
\centering
\includegraphics{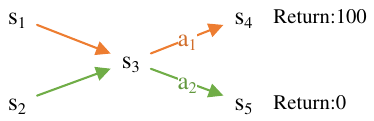}
\caption{Stitching trajectories through conditioning $RTG$ token.}
\label{fig:1}
\end{figure}

Replacing the $RTG$ token from hyperparameters with predicted values can enhance the agent's stitching ability. As shown in Figure \ref{fig:1}, under the assumption of an optimal model aimed at maximizing likelihood, the agent at state $s_3$ selects action $a_1$ when the input $RTG$ is 100, while it opts for action $a_2$ when the input $RTG$ is 0. During the evaluation, although the historical trajectory is $s_2$-$s_3$, it can transition to the state $s_4$ if the input $RTG$ is 100 and ultimately get a higher return. This raises the question of accurately determining an appropriate $RTG$ value. Compared to presetting the $RTG_0$ value, employing a neural network for predicting the $RTG$ token offers two significant advantages. First, the $RTG$ value is expected to be large and appear in the dataset, whereas a presetting $RTG_0$ value cannot adequately meet both criteria simultaneously. Second, the token prediction is state-wise, implying that even if the agent reaches a state with a low expected return due to stochastic transition, the $RTG$ token will not be exceedingly large. To summarize, through the mechanism of token prediction, DT can exhibit excellent performance in stitching ability.

Stitching can be understood as trajectory-level extrapolation. This enhancement allows the agent to generate previously unobserved trajectories but does not enable the agent to select unseen actions. In contrast, state-level extrapolation involves identifying the optimal action at a given state based on knowledge acquired from the dataset without being constrained to the already known actions. To further illustrate state-level extrapolation, we conduct a straightforward experiment to demonstrate the challenge faced by DT in extrapolating beyond their training data.
 
\begin{figure}[t]
\centering
\hspace{-0.45cm}
\includegraphics[width=0.48\textwidth]{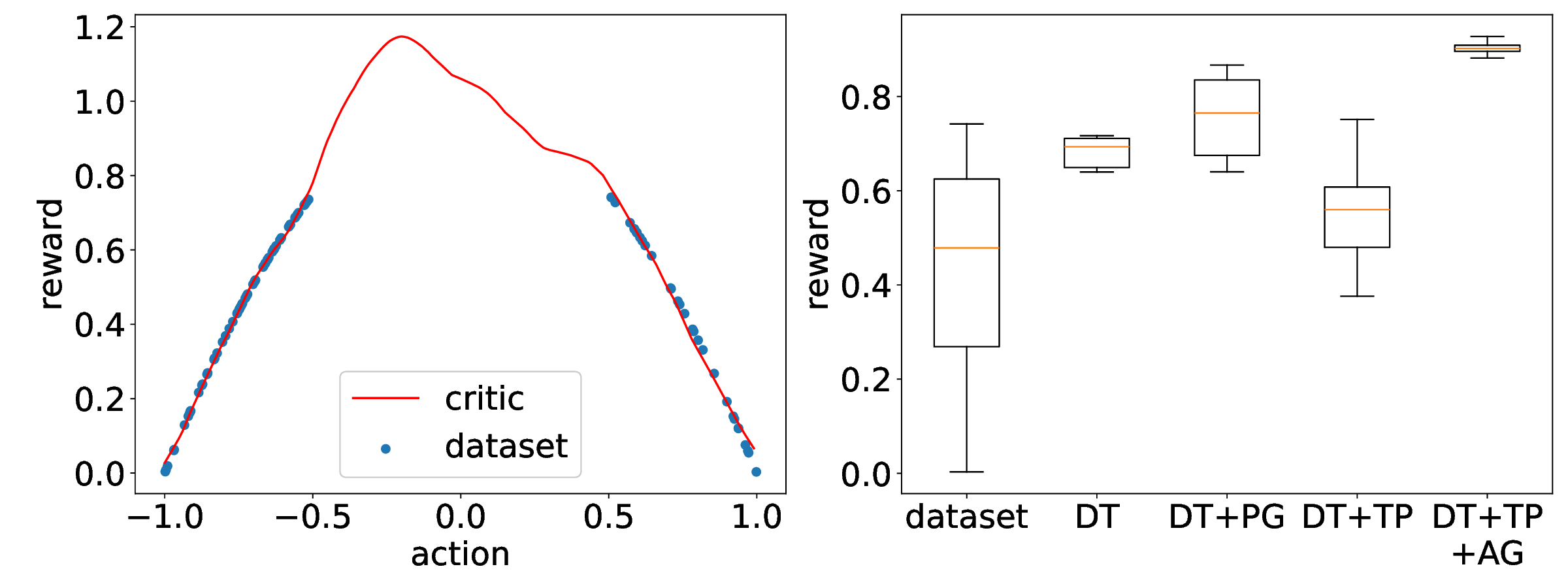}
\caption{The left graph presents the distribution of data within the dataset and the critic's outputs corresponding to various actions. The right graph presents different algorithms' rewards in this special state (DT: Decision Transformer, PG: Policy Gradient, TP: Token prediction, AG: Action Gradient). The state-level extrapolation ability of DT is limited, and token prediction does not effectively address this deficiency. Utilizing a critic to compute gradients can substantially enhance this capability in ways that alternative methods cannot achieve.}
\label{mot:1}
\end{figure}

We establish a simple environment consisting of only one state, where the reward associated with a specific action is defined as $r(a) = 1 - a^2$. In this context, the reward increases as the action approaches zero. However, the dataset utilized for model training is restricted to instances where $|a| > 0.5$, namely, optimal data are absent. In this simple environment, we initially train a critic to approximate the reward function, test its output with various actions, and then implement a range of algorithms, documenting their performance metrics. (see Figure \ref{mot:1}).

In this experiment, the simple reward function can be effectively fitted by a three-layer model, while most practical environments' reward functions are smooth as well. The models using PG or AG are capable of selecting actions that are not present in the training dataset. This straightforward experiment illustrates that state-level extrapolation requires the assistance of the Q-value function and cannot be achieved solely through the TP. 

\subsection{Action Gradient}

Most prior research attempts to use PG to train models with extrapolation abilities; however, our approach truncates the backward propagation process of the policy gradient at the action level. Specifically, the gradient of the Q-value with respect to the parameters of the policy network is not utilized as a loss term during training. Instead, the gradient of the Q-value with respect to the action is computed, and this gradient is backward propagated to adjust the action during the evaluation phase.

The primary modification in our method compared to the original algorithms lies in the evaluation phase. In the training phase, an additional critic needs to be trained for algorithms that do not have a trained critic, which will be discussed in Section \ref{sec:critic}. At each evaluation time step $t$, the following procedure is implemented: firstly, the necessary information is input into the DT to derive the initial action, denoted as $a_t^0$. Subsequently, starting from the initial action, the gradient of the Q-value with respect to the current action is computed. This gradient is added to the current action to derive the following action. The iterative update can be expressed mathematically as:

\begin{align}\label{eq:it_ag}
    a_t^{i+1} = a_t^i + \eta \nabla_{a_t^i} Q(s_t, a_t^i).
\end{align}

\noindent where $\eta$ is a coefficient that controls how much to change the action. The iterative process is repeated for a total of $n$ iterations, resulting in the set of actions $\{a_t^0, a_t^1, \dots, a_t^n\}$. Ultimately, the action with the highest Q-value is selected as the final output, which can be formulated as:

\begin{align}\label{eq:maxq}
    \hat{a_t} = \arg\max_{a_t^i} Q(s_t, a_t^i).
\end{align}

The selected action $\hat{a_t}$ is then employed to interact with the environment, leading to the subsequent state transition. The complete procedure is outlined in Algorithm \ref{alg}.

\begin{algorithm*}[t]
    \caption{Action Gradient}
    \label{alg}
    \begin{algorithmic}
        \REQUIRE Context length $k$, maximum evaluation step $T$, offline dataset $\mathcal{D}$, coefficient $\eta$, iterative times $n$.
        \STATE \textbf{Training}:
        \STATE train the DT policy network $\pi_\theta$ with dataset $\mathcal{D}$.
        \STATE train the critic network $Q_\phi$ with dataset $\mathcal{D}$.
        \STATE \textbf{Inference}:
        \STATE Get initial state $s_0$
        \FOR{$t = 0$ \TO $T$}
            \STATE Predict $RTG_{t}$.
            \STATE $a_t^0 \gets \pi(s_{t-k}, RTG_{t-k}, a_{t-k}, \dots, s_t, RTG_t)$.
            \STATE $S_t \gets \{a_t^0\}$.
            \FOR{$i = 0$ \TO $n-1$}
                \STATE Get $a_t^{i+1}$ by Equation \ref{eq:it_ag} or other improved gradient methods mentioned in Section \ref{sec:ogm}.
                \STATE $S_t \gets S_t \cup \{a_t^i+1\}$.
            \ENDFOR
            \STATE For all the action in set $S_t$, compute their Q-value at state $s_t$.
            \STATE Select action $\hat{a_t}$ by Equation \ref{eq:maxq}.
            \STATE Execute $\hat{a_t}$ and get next state $s_{t+1}$ 
            \IF{$Done$ is $True$}
                \STATE \textbf{break}
            \ENDIF
        \ENDFOR
    \end{algorithmic}
\end{algorithm*}

\subsection{The Method of Critic Training}\label{sec:critic}
In general, we utilize the bootstrap mechanism to train the critic, with the objective of minimizing the Bellman error. The loss function within the context of offline RL is defined as follows:

\begin{align}
    \mathcal{L}_Q=\mathbb{E}_{s,a,r,s^{\prime}\sim \mathcal{D}}\left[\left(r+\gamma\mathbb{E}_{a^{\prime}\sim\pi(s^{\prime})}\left[Q(s^{\prime},a^{\prime})\right] \right.\right. \\ \notag
    \left.\left. -Q(s,a)\right)^2\right].
\end{align}

The rationale for moving away from this naive method is that to remain consistent with the definition of the Bellman operator, the $a^{\prime}$ used when updating the critic is the $a^{\prime}$ corrected by AG. The $a^{\prime}$ is affected by the critic's estimation error, while the error would accumulate during the bootstrapping process, leading to significant overestimation issues. Given this potential issue, it is essential that the critic is trained solely using an offline dataset without the involvement of an agent.

A suitable method that meets the specified requirements is the approach presented in IQL ~\cite{kostrikov2021offline}, which employs expectile regression ~\cite{newey1987asymmetric} to obtain an upper estimation of the Q-value. Considering that trajectories with low returns may cause underestimation, this approach can enhance the accuracy of the Q-value estimation. In this framework, a parameterized V-value function and a parameterized Q-value function are trained by the following loss:

\begin{align}
    \mathcal{L}_{Q_\theta}&=\mathbb{E}_{s,a,r,s^{\prime}\sim \mathcal{D}}\left[\left(r+\gamma V_\phi\left(s^\prime\right)-Q_\theta\left(s,a\right)\right)^2\right], \\
    \mathcal{L}_{V_\phi}&=\mathbb{E}_{s,a\sim \mathcal{D}}\left[\mathcal{L}^\tau_2\left(Q_\theta\left(s,a\right)-V_\phi\left(s\right)\right)\right].
\end{align}

\noindent where $\mathcal{L}^\tau_2(u)=\left|\tau-\mathbbm{1}(u<0)\right|u^2$

\subsection{Other Gradient Method} \label{sec:ogm}

Since the introduction of the backpropagation algorithm ~\cite{rumelhart1986learning}, numerous enhancement techniques have emerged aimed at optimizing neural networks more effectively. Drawing inspiration from these advancements, we seek to incorporate similar improvements into our algorithms. The proposed methods are as follows:

\noindent\textbf{Gradient descent with momentum} ~\cite{qian1999momentum} is a technique designed to mitigate oscillations during the optimization process. By implementing this method, the update equation is modified as follows:

\begin{align}\label{eq:1st}
    v_t^i &= v_t^{i-1} + \zeta\nabla_{a_t^i} Q(s_t, a_t^i), \\ \notag
    a_t^{i+1} &= a_t^i + \eta v_t^i.
\end{align}

\noindent\textbf{Root Mean Square Propagation} ~\cite{hinton2012neural} involves dividing the gradient by the running root mean square, allowing for the adjustment of the learning rate for each parameter individually. This approach assigns reduced learning rates to parameters associated with frequently varying gradients, whereas parameters characterized by infrequent changes are allocated larger learning rates. By implementing this method, the update equation is modified as follows:

\begin{align}\label{eq:2nd}
    g_t^i &= \nabla_{a_t^i} Q(s_t, a_t^i), \\ \notag
    r_t^i &= \zeta r_t^{i-1} + (1-\zeta)(g_t^i)^2, \\ \notag
    a_t^{i+1} &= a_t^i + \frac{\eta}{\sqrt{r_t^i}+\epsilon} g_t^i.
\end{align}

\noindent\textbf{Adaptive Moment Estimation} ~\cite{kingma2017adammethodstochasticoptimization} is a combination of two concepts: momentum and RMSProp. This combination allows Adam to effectively adapt the learning rate for each parameter, providing benefits from both momentum and RMSProp techniques. By implementing this method, the update equation is modified as follows:

\begin{align}\label{eq:adam}
    g_t^i &= \nabla_{a_t^i} Q(s_t, a_t^i), \\ \notag
    m_t^i &= \zeta_1 m_t^{i-1} + (1-\zeta_1)g_t^i,\quad \hat{m_t^i}=\frac{m_t^i}{1-\zeta_1} \\ \notag
    v_t^i &= \zeta_2 v_t^{i-1} + (1-\zeta_2)(g_t^i)^2,\quad \hat{v_t^i}=\frac{v_t^i}{1-\zeta_2} \\ \notag
    a_t^{i+1} &= a_t^i + \frac{\eta}{\sqrt{v_t^i}+\epsilon} m_t^i.
\end{align}

Incorporating these three improvements separately, we conduct experiments on AG. In some environments, we observed performance improvements. More details can be found in the Section \ref{sec:abla}. There is no doubt that these improvements can play their intended role, but their effectiveness may be interfered with by the errors present in the critics.

\section{Experiment}

We conducted a series of experiments to address the following research questions: First, to what extent does the application of AG enhance the performance of DT algorithms? Second, how does AG compare to PG in terms of effectiveness when integrated with token prediction? Third, what is the impact of related hyperparameters and gradient methods on overall performance? 

\subsection{Benchmarks and Baseline Algorithms}

The experiments are conducted on the widely recognized D4RL datasets ~\cite{fu2020d4rl}, including locomotion and navigation tasks. The basic algorithm we chose is Reinformer (RF) ~\cite{zhuang2024reinformer}. This choice is predicated on the fact that RF, by separating the RTG prediction network, minimally alters the training process relative to the original DT. Specifically, RF employs NLL loss rather than MSE loss during training. Furthermore, it leverages the token prediction technique during the evaluation phase, resulting in superior performance. After incorporating AG, we evaluate its performance against traditional algorithms, including BC ~\cite{pomerleau1988alvinn}, TD3+BC ~\cite{fujimoto2021minimalist}, CQL ~\cite{kumar2020conservative}and IQL ~\cite{kostrikov2021offline}, as well as DT-based algorithms, including DT ~\cite{chen2021decision}, CGDT ~\cite{wang2024critic}, ADT ~\cite{ma2023rethinking} and original RF. Except for the results about CGDT, ADT, and RF, which are from their original study, the remaining results are sourced from CORL ~\cite{tarasov2024corl}. Our experimental results are obtained by testing the performance of five distinct random seeds, calculating the average after evaluating the model for ten episodes with each seed.

The implementation of RF ~\cite{zhuang2024reinformer} is based on the original paper, but there are some subtle differences. Firstly, the original paper used an identical network for predicting RTG and action, whereas we have separated them into two independent networks. This change has minimal impact on the algorithm's performance but slightly improves its stability. Secondly, while the original paper employed different hyperparameters for different environments, our experiments used a unified set of hyperparameters across all environments. Considering the impact of context length on the algorithm, we designed an adaptive context length mechanism inspired by EDT ~\cite{wu2024elastic}: during the evaluation phase, when the previous RTG is greater than the current RTG, it is included as input to the algorithm until either this condition is no longer met or an upper limit is reached.

\subsection{Main Results}

The results of RF with AG and other baseline algorithms are presented in Table \ref{tab:result}. With the exception of the maze2d-large environment, performance improvements were observed across all tested environments. While our algorithm attains the highest scores in only a subset of these environments, the overall performance surpasses that of the baseline algorithms. These results demonstrate that, by combining AG and advanced token prediction techniques, the algorithm can significantly outperform prior DT-based algorithms. 

\begin{table*}[t]
    \centering
    \begin{tabular}{l|cccc|ccccc}
        \bhline
        \multicolumn{1}{c|}{\textbf{Environment}} & \textbf{BC} & \textbf{TD3+BC} & \textbf{CQL} & \textbf{IQL} & \textbf{DT} & \textbf{CGDT} & \textbf{ADT} & \textbf{RF} & \textbf{RF+AG} \\
        \hline
        halfcheetah-medium & 42.4 & 48.1 & 47.0 & 48.3 & 42.2 & 43.0 & \textbf{48.7} & 42.9 & 46.1$\pm$0.3 \\
        halfcheetah-medium-replay & 35.7 & 44.8 & 45.0 & 44.5 & 38.9 & 40.4 & \textbf{42.8} & 39.0 & 42.4$\pm$0.2 \\
        halfcheetah-medium-expert & 56.0 & 90.8 & 95.6 & 94.7 & 91.6 & \textbf{93.6} & 91.7 & 92.0 & 92.3$\pm$0.4 \\
        hopper-medium & 53.5 & 60.4 & 59.1 & 67.5 & 65.1 & 96.9 & 60.6 & 81.6 & \textbf{98.9$\pm$0.8} \\
        hopper-medium-replay & 29.8 & 64.4 & 95.1 & 97.4 & 81.8 & \textbf{93.4} & 83.5 & 83.3 & 91.4$\pm$3.7 \\
        hopper-medium-expert & 52.3 & 101.2 & 99.3 & 107.4 & 110.4 & 107.6 & 101.6 & 107.8 & \textbf{111.0$\pm$0.7} \\
        walker2d-medium & 63.2 & 82.7 & 80.8 & 80.9 & 67.6 & 79.1 & 80.9 & 80.5 & \textbf{86.0$\pm$1.3} \\
        walker2d-medium-replay & 21.8 & 85.6 & 73.1 & 82.2 & 59.9 & 78.1 & \textbf{86.3} & 72.9 & 79.3$\pm$1.5 \\
        walker2d-medium-expert & 99.0 & 110.0 & 109.6 & 111.7 & 107.1 & 109.3 & \textbf{112.1} & 109.4 & 110.4$\pm$0.4 \\
        \hline
        gym-total & 453.7 & 688.0 & 704.6 & 734.6 & 664.6 & 741.4 & 708.2 & 709.4 & \textbf{757.8} \\
        \hline
        maze2d-umaze & 0.4 & 29.4 & -8.9 & 42.1 & 18.1 & / & / & 57.2 & \textbf{71.5$\pm$9.6} \\
        maze2d-medium & 0.8 & 59.5 & 86.1 & 34.9 & 31.7 & / & / & 85.6 & \textbf{90.2$\pm$5.1} \\
        maze2d-large & 2.3 & 97.1 & 23.8 & 61.7 & 35.7 & / & / & \textbf{47.4} & 32.2$\pm$4.5 \\
        \hline
        maze2d-total & 3.5 & 186.0 & 101.0 & 138.7 & 85.5 & / & / & 190.2 & \textbf{193.9} \\
        \bhline
    \end{tabular}
    \caption{The normalized scores of Reinformer with AG (RF+AG) and other baseline algorithms. Traditional algorithms and DT-based algorithms are separated to the left and right sides. The best scores among all DT-based algorithms are \textbf{bold}.}
    \label{tab:result}
\end{table*}

\subsection{Comparison Experiments with Policy Gradient}\label{sec:compexp}

We claim that the mere incorporation of policy gradient techniques into DT-based methods does not lead to a stable improvement in performance. To validate this claim, based on RF, we introduce a policy gradient term to the loss function, drawing upon RF, and assess its efficacy. We examine three distinct methodologies that utilize a critic network: PG, AWAC ~\cite{nair2020awac}, and AG. To ensure the internal validity of our experiment, the implementation of these three methods adheres to the OAMPI paradigm ~\cite{brandfonbrener2021offline}, employing critic networks that share identical parameters and architecture. The implementation of PG references the work of QT ~\cite{hu2024q}, which involves the addition of a normalized policy gradient term to the loss function. The loss function is:

\begin{align}
    \mathcal{L}_{\pi}=\mathcal{L}_{DT}-\alpha\frac{\mathbb{E}_{\substack{\tau_t\sim\mathcal{D},\\s_i\sim\tau_t}}\left[Q(s_i,\pi(\tau_t)_i)\right]}{\mathbb{E}_{\substack{\tau_t\sim\mathcal{D},\\s_i\sim\tau_t}}\left[|Q(s_i,\pi(\tau_t)_i)|\right]}.
\end{align}

\noindent The AWAC method is another method that uses a critic in the training phase. The AWR ~\cite{wang2018exponentially,peng2019advantage} method, which is similar to the AWAC method, is used in the ADT ~\cite{ma2023rethinking} method, and is not chosen for our experiments since the AWR method also involves estimation of the V-value. This method does not add an extra term but instead adds weight to the original loss, which is defined as follows:

\begin{align}
    \mathcal{L}_{\pi}= \mathbb{E}_{\substack{\tau_t\sim\mathcal{D},\\s_i,a_i\sim\tau_t}}
    \bigg[ &\text{log}p(a_i|\pi(\tau_t)_i)\times  \\ \notag
    &\text{exp}(\frac{1}{\lambda}(Q(s_i,a_i)-Q(s_i,\pi(\tau_t)_i))) \bigg].
\end{align}

\noindent It is noticeable that this loss involves the NLL loss term rather than the MSE loss term. This approach diverges from the original DT ~\cite{chen2021decision} but aligns with several improved algorithms ~\cite{zheng2022online,zhuang2024reinformer,yan2024reinforcement}. Given that our baseline algorithm is RF, no modifications to the implementation are necessary, and the entropy loss term is computed independently. Except for the loss function (AG is not modified, but it modifies the evaluation step), the hyperparameters are kept the same, and the results are displayed in Table \ref{tab:compwpg}. It is reasonable that, in a majority of environments, the incorporation of a critic network, regardless of the method employed, improves the performance of the algorithm. As discussed in Section \ref{sec:extrapolation}, the ability for state-level extrapolation is augmented by these methods. On the other hand, from AWAC's theory, moves with high Q in the trajectory are given higher weights, and PG would have a similar effect.

\begin{table*}[t]
    \centering
    \begin{tabular}{c|c|c|c|c}
        \bhline
        \textbf{Environment} & \textbf{RF} & \textbf{RF+PG} & \textbf{RF+AWAC} & \textbf{RF+AG} \\
        \hline
        halfcheetah-medium & 42.9$\pm$0.4 & 43.3$\pm$0.3(+0.9\%) & 46.6$\pm$0.2(+8.6\%) & 46.1$\pm$0.3(+7.5\%) \\
        \hline
        hopper-medium & 81.6$\pm$3.3 & 96.2$\pm$2.0(+17.9\%) & 87.7$\pm$18.8(+7.5\%) & 98.9$\pm$0.8(+21.2\%) \\
        \hline
        walker2d-medium & 80.5$\pm$2.7 & 78.6$\pm$2.6(-2.4\%) & 79.0$\pm$2.9(-1.9\%) & 86.0$\pm$1.3(+6.8\%) \\
        \hline
        halfcheetah-medium-expert & 92.0$\pm$0.3 & 91.8$\pm$0.2(-0.2\%) & 51.4$\pm$4.6(-44.1\%) & 92.3$\pm$0.4(+0.3\%) \\
        \hline
        hopper-medium-expert & 107.8$\pm$2.1 & 102.7$\pm$1.8(-4.7\%) & 30.1$\pm$25.3(-72.0\%) & 111.0$\pm$0.7(+3.0\%) \\
        \hline
        walker2d-medium-expert & 109.4$\pm$0.3 & 109.4$\pm$0.2(+0\%) & 107.1$\pm$6.2(-2.1\%) & 110.4$\pm$0.4(+0.9\%) \\
        \bhline
    \end{tabular}
    \caption{The normalized scores of naive RF, RF+PG, RF+AWAC, RF+AG. In most environments, no matter the method used, the introduction of the critic network enhances the algorithm's performance.}
    \label{tab:compwpg}
\end{table*}

These empirical results do not provide conclusive evidence that AG significantly outperforms PG, AWAC, and similar methods. On the one hand, there is a diverse range of techniques available for training a critic and applying the gradient of the Q-value. On the other hand, the consistency of our experimental setup imposes limitations on the performance of these methods, suggesting that enhancements could be achieved through hyperparameter tuning and the integration of additional techniques. Nonetheless, our results indicate that incorporating a critic during the evaluation rather than the training phase can also improve algorithm performance. This approach capitalizes on the advantages of AG discussed in Section \ref{sec:comparison} and offers new insights for future research.

\subsection{Ablation Experiments}\label{sec:abla}

\noindent\textbf{Ablation on Token Prediction} Since RF is a relatively pure algorithm applying the TP technique, we do not conduct ablation experiments on other algorithms using the TP technique due to potential interference. However, we do experiments on naive DT to investigate the performance of the AG method in the absence of the TP technique, a condition in which the stitching capability may be diminished. The results are presented in Table \ref{tab:compondt}. Overall, there is a notable performance improvement; however, in specific environments, the enhancement is minimal or, in some cases, even slightly diminished. The limited effectiveness of AG can be attributed to the scarcity of stitching ability, which forces the agent to follow the existing trajectory even when enhanced actions are generated. Consequently, even when the agent generates enhanced actions aiming at reaching states with high expected returns, it is compelled to revert to the original trajectory.

\begin{table}[ht]
    \centering
    \begin{tabular}{c|c|c}
        \bhline
        \textbf{Environment} & \textbf{DT} & \textbf{DT+AG} \\
        \hline
        halfcheetah-medium & 42.2$\pm$0.3 & 41.7$\pm$0.2 \\
        \hline
        hopper-medium & 65.1$\pm$1.6 & 66.7$\pm$3.1 \\
        \hline
        walker2d-medium & 67.6$\pm$2.5 & 76.9$\pm$2.1 \\
        \hline
        halfcheetah-medium-replay & 38.9$\pm$0.5 & 39.8$\pm$0.7 \\
        \hline
        hopper-medium-replay & 81.8$\pm$6.8 & 87.3$\pm$2.2 \\
        \hline
        walker2d-medium-replay & 59.9$\pm$2.7 & 74.3$\pm$4.0\\
        \bhline
    \end{tabular}
    \caption{Comparison between naive DT and DT+AG. }
    \label{tab:compondt}
\end{table}

\begin{figure}[ht]
\centering
\includegraphics[width=0.48\textwidth]{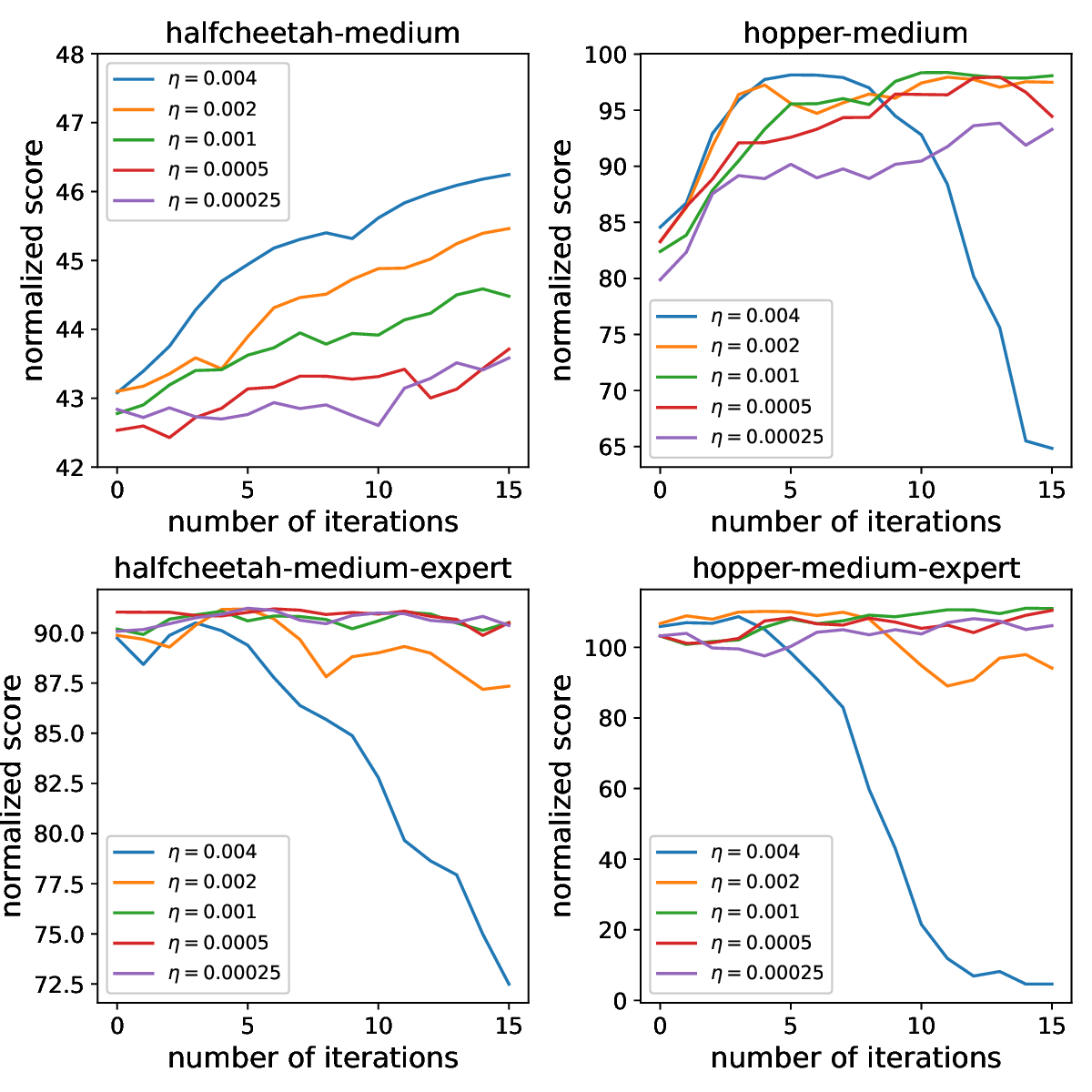}
\caption{The experimental results of the ablation study focusing on the coefficient $\eta$ and the number of iterations $n$.}
\label{fig:abla}
\end{figure}

\noindent\textbf{Ablation on Hyperparameters} Next, we conduct ablation experiments to explore the effectiveness of the coefficient $\eta$ and the number of iterations $n$. The results are illustrated in Figure \ref{fig:abla}. According to the definition, AG is not applied when $n=0$. Except for the halfcheetah-medium environment, a larger value of $\eta$ leads to a deterioration in performance even after several iterations. It is evident in the two medium environments that as the number of iterations $n$ increases, the score improves; furthermore, a higher $\eta$ is associated with a more rapid rate of increase.

\noindent\textbf{Ablation on Gradient Method} As elaborated in Section \ref{sec:ogm}, referencing the gradient descent optimization algorithms, we propose the integration of first-order and second-order momentum into AG. Although the incorporation of momentum can enhance the performance of algorithms in certain environments, the results presented in Table \ref{tab:abgm} do not provide sufficient evidence to evaluate the effectiveness of various gradient methods. We consider that some environments require crossing the optimal point along the direction of the gradient, due to the fact that the true optimal point and the estimated optimal point may not coincide. This leads to poor performance of these gradient methods. A more comprehensive experimental approach, incorporating hyperparameter tuning, is necessary to investigate and identify superior gradient methods further.

\begin{table*}[t]
    \centering
    \begin{tabular}{c|c|c|c|c}
        \bhline
        \textbf{Environment} & \textbf{None} & \textbf{FM} & \textbf{SM} & \textbf{FM+SM} \\
        \hline
        halfcheetah-medium & 46.1$\pm$0.3 & 44.8$\pm$0.3 & 44.7$\pm$0.1 & 45.9$\pm$0.7 \\
        \hline
        hopper-medium & 98.2$\pm$0.6 & 98.4$\pm$0.1 & 89.8$\pm$6.8 & 98.9$\pm$0.8 \\
        \hline
        walker2d-medium & 86.0$\pm$1.3 & 81.8$\pm$0.9 & 78.8$\pm$2.7 & 84.0$\pm$1.5 \\
        \hline
        halfcheetah-medium-expert & 91.2$\pm$0.5 & 70.5$\pm$9.9 & 90.5$\pm$0.8 & 92.3$\pm$0.4 \\
        \hline
        hopper-medium-expert & 111.0$\pm$0.7 & 109.6$\pm$0.9 & 108.5$\pm$4.2 & 99.4$\pm$7.2 \\
        \hline
        walker2d-medium-expert & 110.4$\pm$0.4 & 109.4$\pm$0.3 & 109.3$\pm$0.4 & 110.3$\pm$0.3 \\
        \bhline
    \end{tabular}
    \caption{The normalized score of algorithms applying different gradient methods. None means no moments are used (See Equation \ref{eq:it_ag}). FM introduces the first moment (See Equation \ref{eq:1st}), SM introduces the second moment See Equation \ref{eq:2nd}), and FM+SM introduces both the first and second moments simultaneously (See Equation \ref{eq:adam}).}
    \label{tab:abgm}
\end{table*}

\section{Discussion}\label{sec:comparison}

Incorporating a policy gradient term into the loss function has been widely acknowledged as an effective approach to enhance DT-based methods ~\cite{hu2024q,yan2024reinforcement}. This raises a question regarding the value of exploring alternative methods. In this section, we will discuss the limitations of PG and the advantages offered by AG. The fundamental difference between AG and PG (or other methods using Q-value function during training phase ~\cite{wang2024critic,ma2023rethinking}) is that AG is an independent module, which means regardless of whether improvements from the evaluation phase or the training phase are applied based on DT, AG only functions after the model outputs actions. This independence brings two benefits: compatibility and convenience for hyperparameter optimization.

\noindent\textbf{Compatibility} The existing DT with PG algorithm, as discussed in ~\cite{hu2024q}, does not represent a universal enhancement; its superior performance cannot be solely attributed to the integration of policy gradient term. The experiment based on Reinformer ~\cite{zhuang2024reinformer} can demonstrate that merely using PG is insufficient (see \ref{sec:compexp}). It is important to note that Reinformer does not utilize a Q-value function during its training phase. This characteristic facilitates straightforward integration with PG, resulting in performance improvements that are consistent with the analysis presented in Section \ref{sec:extrapolation}. In contrast, determining an appropriate loss function is more complex if original algorithms incorporate a Q-value function into their loss function ~\cite{wang2024critic,ma2023rethinking}. Conversely, AG can be easily integrated into various algorithms for two primary reasons. First, there is no fundamental conflict between these enhanced algorithms and AG. Second, avoiding the propagation of Q-value estimation errors can improve numerical stability. The integration of AG is unlikely to compromise the performance of the original algorithms, as it does not interfere with their stabilization. This is a critical factor in offline RL, where multiple algorithms often maintain a delicate equilibrium due to the accumulation of various errors.

\noindent\textbf{Hyperparameter Optimization} It is well-known that numerous RL algorithms involve a significant number of hyperparameters. This issue is particularly pronounced in the context of offline RL, where many algorithms introduce additional hyperparameters. PG is not exempt from this challenge; specifically, the coefficients that govern the balance between the original term and the policy gradient term necessitate extensive experimentation for optimal selection. While AG incorporates extra hyperparameters as well, tuning these parameters requires only a reiteration of the evaluation phase since this method does not alter the training process. We advocate for a hyperparameter optimization strategy that first involves selecting optimal hyperparameters for the basic model, then applying AG, and finally, tuning the associated hyperparameters. 

In future research endeavors, investigations into DT-based algorithms can concentrate on augmenting trajectory-level extrapolation abilities. Concurrently, improvements in state-level extrapolation can be achieved by developing enhanced AG methods. By delineating the extrapolation challenge into distinct components, the design of DT-based algorithms can become more focused and effective. We posit that the exploration of advanced gradient methods and optimized critic training techniques have the potential to enhance AG's performance. Furthermore, the refinement of token prediction methods could substantially improve the stitching abilities of DT-based algorithms. The integration of these enhancements is expected to produce robust and comprehensive algorithms. Therefore, our objective is to apply AG to develop foundational techniques that will expand the possibilities for future DT-based algorithms.

\section{Conclusion}

In this work, we propose the Action Gradient (AG) method, which has significant potential in addressing the extrapolation challenges present in DT-based algorithms. Experimental results based on multiple tests using the D4RL benchmark dataset show that the algorithm integrating Token Prediction (TP) and AG outperforms prior DT-based algorithms in various environments, validating its effectiveness. These findings offer new perspectives and methodologies for algorithm design in the domain of offline RL. Future research can concentrate on the optimization of AG and the exploration of the integration with other advanced TP techniques to improve performance and advance the application of offline RL across broader domains.

\clearpage

\bibliographystyle{named}
\bibliography{ijcai25}

\end{document}